\documentclass[letterpaper,10pt,conference]{ieeeconf}  % Comment this line out
\IEEEoverridecommandlockouts                              % This command is only
\usepackage{graphicx} % for pdf, bitmapped graphics files
\usepackage{amsmath} % assumes amsmath package installed
\usepackage{amssymb}  % assumes amsmath package installed
\usepackage{algorithm}
\usepackage{algpseudocode}
\usepackage{xcolor}
\makeatletter
\let\NAT@parse\undefined
\makeatother
\usepackage{hyperref}
\usepackage{subfig}
\usepackage{float}
\graphicspath{{figures/}}

\title{\LARGE \bf Functional Task Tree Generation from a Knowledge Graph to Solve Unseen Problems}

\author{Md. Sadman Sakib, David Paulius, and Yu Sun\\% <-this % stops a space\\
\thanks{
Yu Sun and Md. Sadman Sakib are members of the Robot Perception and Action Lab (RPAL), which is part of the Department of Computer Science \& Engineering at the University of South Florida, Tampa, FL, USA. David Paulius is a postdoctoral researcher in the Human-centered Assistive Robotics (HCR) group at the Technical University of Munich, Germany.
\newline\textit{Email}: \texttt{\{mdsadman,yusun\}@usf.edu, david.paulius@tum.de}}%
}

\newcommand{\TaskTree}{\mathcal{T}}
\newcommand{\Subtree}{\mathcal{T_S}}
\newcommand{\Ingredients}{\mathbb{I}}
\newcommand{\DishType}{\mathsf{D}}
\newcommand{\Candidates}{\mathsf{candidates}}
\newcommand{\Subgraph}{\mathsf{subgraph}}
\newcommand{\Goal}{\mathsf{goal}}
\newcommand{\G}{\mathbf{G}}
\newcommand{\SetS}{\mathbb{S}}

\newcommand{\SetT}{\mathbb{T}}
\newcommand{\Score}{\mathsf{score}}
\newcommand{\Count}{\mathsf{count}}
\newcommand{\FU}{\mathsf{FU}}

\newcommand{\FOON}{\mathsf{FOON}}

\begin{document}

\maketitle

\thispagestyle{empty}
\pagestyle{empty}

%%%%%%%%%%%%%%%%%%%%%%%%%%%%%%%%%%%%%%%%%%%%%%%%%%%%%%%%%%%%%%%%%%%%%%%%%%%%%%%%
\begin{abstract}

A major component for developing intelligent and autonomous robots is a suitable knowledge representation, from which a robot can acquire knowledge about its actions or world. However, unlike humans, robots cannot creatively adapt to novel scenarios, as their knowledge and environment are rigidly defined. To address the problem of producing novel and flexible task plans called \textit{task trees}, we explore how we can derive plans with concepts not originally in the robot’s knowledge base. Existing knowledge in the form of a knowledge graph is used as a base of reference to create task trees that are modified with new object or state combinations. To demonstrate the flexibility of our method, we randomly selected recipes from the Recipe1M+ dataset and generated their task trees. The task trees were then thoroughly checked with a visualization tool that portrays how each ingredient changes with each action to produce the desired meal. Our results indicate that the proposed method can produce task plans with high accuracy even for never-before-seen ingredient combinations.

\end{abstract}
%%%%%%%%%%%%%%%%%%%%%%%%%%%%%%%%%%%%%%%%%%%%%%%%%%%%%%%%%%%%%%%%%%%%%%%%%%%%%%%%

\section{Introduction}

An ongoing trend for research in robotics is the development of robots that can jointly understand human intentions and actions and execute manipulations for human problem domains and tasks \cite{fong2003survey, chandrasekaran2015human, goodrich2008human}. 
Some of the interesting applications include assistance for the elderly and disabled person \cite{DBLP:conf/iros/WadaSST03, DBLP:conf/iros/HarmoTKVH05, DBLP:journals/ar/WuWCW21}, cleaning \cite{DBLP:journals/ras/MitalKHA97, DBLP:journals/access/LiXT21}, food delivery \cite{DBLP:conf/icra/XueRHKD11, DBLP:journals/ral/LinWCHC21} etc. 
A key component needed for intelligent and autonomous robots is a knowledge representation~\cite{paulius2019survey}, which would allow robots to understand its actions in a way that is also interpretable by humans.
Inspired by this goal and previous work on joint object-action representation~\cite{Ren2013,SunRAS2013,Lin2015a}, we introduced the \textit{functional object-oriented network} (FOON) as a knowledge representation for service robots~\cite{paulius2016functional,paulius2018functional}.
FOON, which is motivated by the theory of affordance \cite{Gibson_1977}, takes the form of a bipartite graph to describe the relationship between objects and manipulation actions as nodes in the network.
We described how knowledge can be combined into a single source from which knowledge can be retrieved as task sequences called task trees~\cite{paulius2016functional}, while in~\cite{paulius2018functional}, we demonstrated how existing knowledge can be used to learn ``new'' concepts based on semantic similarity. 

\begin{figure}[t]
	\centering
	\includegraphics[width=\columnwidth]{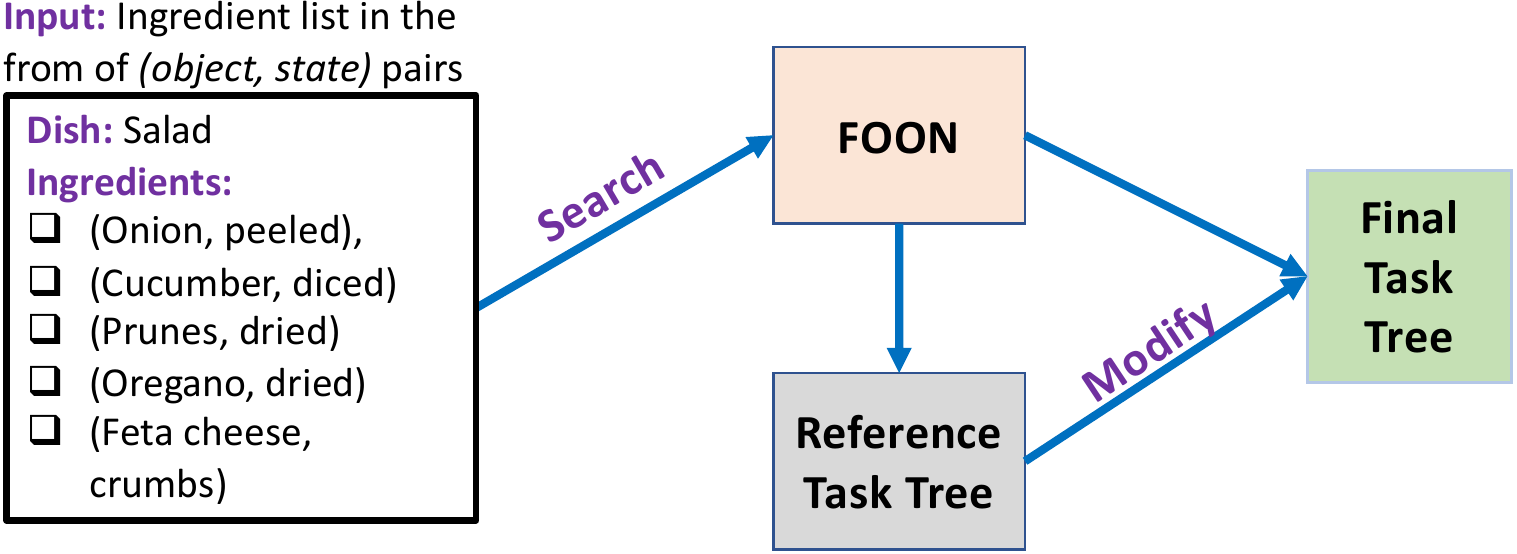}
	\caption{Overview of our task tree generation procedure for novel problems using FOON. As input to our framework, a set of ingredients and their states are given, from which a novel task tree can be generated.}
	\label{fig:pipeline}
\end{figure}

% In this work, we build upon our previous work of using semantic similarity to adapt our FOON to unseen problems for which it does not have complete knowledge to solve. Previously, we assumed that a FOON will contain all of the knowledge for the robot's domain and environment; however, this is not always the case in the real world due to how dynamic problems and environments may be.
% For instance, when considering the problem domain of cooking, a robot will have the knowledge to prepare a specific recipe, but there may be times when this recipe needs to be adapted by including other ingredients or items.
% Humans exhibit this sense of creativity when solving problems, where we can dynamically adapt a recipe or procedure and infer missing information to complete the task.
% Similarly, ingredients may or may not exist in the robot's knowledge base, but the robot may infer how it can use or manipulate certain objects if it is similar to something it already knows about~\cite{paulius2018functional}. 

Previously, we assumed that a FOON will contain all of the knowledge for the robot's domain and environment; however, this is not always the case in the real world due to how dynamic problems and environments may be.
For instance, when considering the problem domain of cooking, a robot will have the knowledge to prepare a specific recipe, but there may be times when this recipe needs to be adapted by including other ingredients or items.
Humans exhibit this sense of creativity when solving problems, where we can dynamically adapt a recipe or procedure and infer missing information to complete the task.
Similarly, ingredients may or may not exist in the robot's knowledge base, but the robot may infer how it can use or manipulate certain objects if it is similar to something it already knows about~\cite{paulius2018functional}. In this work, we propose a solution based on  semantic similarity of ingredients to adapt our FOON to unseen problems for which it does not have complete knowledge to solve. 
 
Figure \ref{fig:pipeline} presents an overview of the task tree generation procedure discussed in this paper. The input is a list of ingredients in the form of \textit{(object, state)} pairs, and the output is a task tree that a robot can follow to prepare the dish. First, we identify a \textit{reference goal object node} and generate a \textit{reference task tree} using FOON. Then, we modify it using the concept of semantic similarity to generate the final task tree. 
In our experiments, we use the Recipe1M+ dataset~\cite{marin2019learning} as a source of information from which we can produce task trees.
Our contributions in this paper are as follows:
\begin{itemize}
    \item We propose a scheme to generate recipes using a knowledge graph for any combination of ingredients. The time and space complexity of our search is optimized by pruning irrelevant branches while traversing the graph.
    
    \item We show how to derive task plan for unseen items using semantic similarity with other existing items. The quality of generated task plans can improve through the addition of diverse set of ingredients to FOON.
    
    \item We introduce a method to evaluate the quality of a task plan represented as a task tree. Our visualization tools help us to quickly identify mistakes in the task tree. 
    
\end{itemize}

% This paper is structured as follows: first, we briefly review the FOON representation in Section~\ref{sec:FOON}. In Section~\ref{sec:task_tree_mod}, we then introduce the task tree generation problem and discuss how this is done. 
% Afterward, we discuss our experiments, findings, and conclusions in Sections~\ref{sec:experiments} and \ref{sec:con}.

% {\color{red}[Fill in more related works or problems... need to elaborate on novelty of the problem we are addressing.]}

% \begin{figure}[t]
% 	\centering
% 	\includegraphics[width=\columnwidth]{fig-universal_FOON-2.pdf}
% 	\caption{Illustration of a universal FOON made of 111 instructional videos. 
% 	This network, along with other subgraphs, are available on our website \cite{foonet}.}
% 	\label{fig:FOON}
% \end{figure}

% Our paper is organized as follows: in Section \ref{sec:FOON}, we give a short overview of the FOON structure, which is followed by Section \ref{sec:retr} where we review the idea behind task tree retrieval.
% In Section \ref{sec:task_tree_mod}, we introduce the problem of task tree generation and modification based on an unseen set of ingredients.
% Finally, in Section \ref{sec:exp}, we discuss our evaluations of our pipeline by assessing the quality of attained task trees.

\section{Functional Object-Oriented Network}
\label{sec:FOON}
% David: fill in the background information here, talk about what has been done in FOON so far.
% {\color{blue}[C-1.2] 
% The functional object-oriented network represents manipulations as seen in cooking activities by capturing the objects and the activity's motions as a graphical structure.
% Previously, FOON was introduced as a knowledge graph-based representation through which high-level concepts related to human manipulations can be represented for service robots.

% Object nodes symbolize any object that is manipulated passively or actively within the activities in FOON, while motion nodes symbolize the type of manipulation that object nodes are participating in a given task.
% These motion nodes can be actions performed in cooking such as pouring, cutting, or stirring, but they can also be extended to manipulations in other domains.

% }
% The joint representation of object and motion nodes.
% As with typical bipartite networks, object nodes can only connect to motion nodes, and motion nodes can only connect to object nodes.

\subsection{Basics of FOON}
\begin{figure}[t]
	\centering
    \includegraphics[width=6cm]{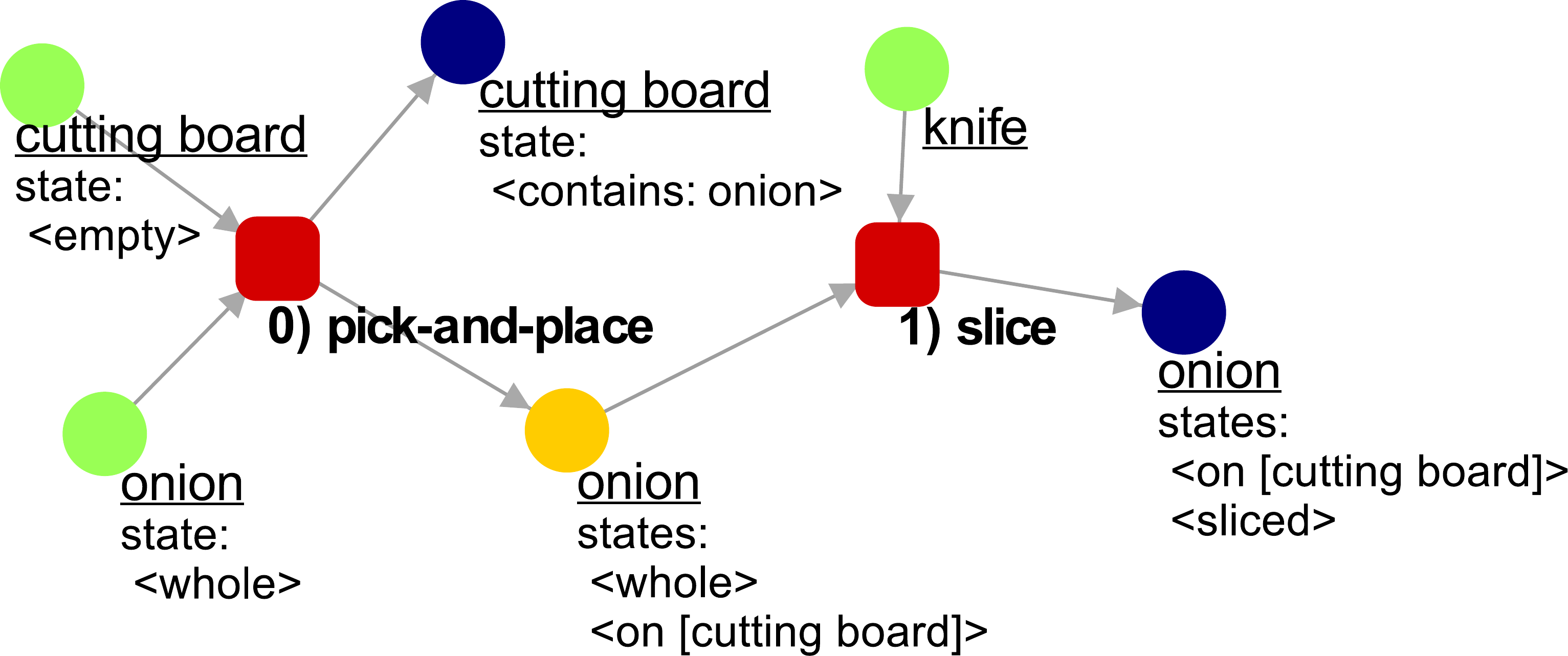}
	\caption{An example of two connected functional units (best viewed in color) with three input nodes (in green), an intermediary node (in yellow), and two output nodes (in dark blue) connected via a motion node (in red), which describe placing an onion on a cutting board and then slicing it.
	}
	\label{fig:unit}
\end{figure}

A typical FOON graph has two types of nodes, \textit{object nodes} and \textit{motion nodes}, which makes it a bipartite network.
Affordances~\cite{Gibson_1977} are depicted with edges that connect objects to actions, which also indicate order of actions in the network.
To represent actions in FOON, we use a collection of object and motion nodes to describe a single action within an activity called a \textit{functional unit}.
A functional unit describes the state change of objects before and after execution, where states can be used to determine when an action has been completed~\cite{jelodar2018identifying,jelodar2019joint}.
% The coupling of object and motion nodes to represent a single action is referred to as a \textit{functional unit}.
% {\color{red}[C-2.2] 
{\it Input} object nodes describe the required state(s) of objects for an action, and {\it output} object nodes describe the outcome of executing said action.
% }
In Figure~\ref{fig:unit}, we show an example of two functional units for: 1) picking-and-placing an onion onto a cutting board, and 2) slicing the onion with a knife.
Some actions may not cause a change in all input objects' states, so there may be instances where there are fewer output object nodes than inputs; for instance, in Figure~\ref{fig:unit}, the \textit{onion} object's state changes across each functional unit, while the \textit{knife} object does not change as a result of the slicing action.
% Each functional unit contains a single motion node describing the action.
% {\color{red}[C-2.1] 

% \definecolor{officegreen}{rgb}{0.0, 0.5, 0.0}

% {\color{red}[C-2.6] 
A FOON is created by annotating video demonstrations into the FOON graph structure.
% ; in this annotation process, we note the actions, objects, and state changes (as functional units) that eventually result in a specific meal or product.
% Typically, an activity is represented by a sequence of functional units that are connected by common object nodes.
% }
% {\color{officegreen}[C-1.1]/ [C-2.3] / [C-2.6]
A FOON that represents a single activity is referred to as a {\it subgraph}; a subgraph contains functional units in sequence to describe objects' states before and after each action occurs,  and what objects are being manipulated.
Presently, annotation is done manually, but efforts have been made to investigate how we can annotate graphs in a semi-automatic procedure~\cite{jelodar2018long}.
Two or more subgraphs can be merged together to form what we call a {\it universal FOON}. 
A universal FOON could propose variations of recipes once it has been created from merging graphs obtained from several sources of knowledge. 
This merging procedure is simply a union operation applied to all functional units from each subgraph we wish to combine; as a result, duplicate functional units are eliminated.
Presently, our FOON dataset comprises of 111 subgraph annotations of videos from YouTube, Activity-Net~\cite{caba2015activitynet}, and EPIC-KITCHENS~\cite{Damen2018EPICKITCHENS}, which are all available on our website~\cite{foonet}.

\subsection{Task Planning with FOON}
In addition to representing knowledge, a robot can use FOON for problem solving.
% {\color{red}[C-2.3] 
Through a process of \textit{task tree retrieval}, a robot can obtain a subgraph, known as a \textit{task tree}, that contains the steps (i.e., functional units) it needs to execute to achieve some sort of goal.
% A subgraph that is obtained from knowledge retrieval is called a {\it task tree}.
A task tree differs from a regular subgraph, as it may not exactly match  human demonstration; rather, it will leverage knowledge from multiple sources to produce a novel task sequence.
The conventional retrieval algorithm requires a list of items in its environment to identify ideal functional units based on the availability of inputs to these units~\cite{paulius2016functional}.
The algorithm draws upon the depth-first search (DFS) and breadth-first search (BFS) algorithms: starting from the goal node, we search for candidate functional units in a depth-wise manner, while for each candidate unit, we search among its input nodes in a breadth-wise manner to determine whether or not they are in the kitchen.
In~\cite{paulius2021weighted}, we introduced an alternative retrieval algorithm that explores different paths to achieving a goal; this was used to evaluate the most optimal path with a weighted FOON, where each functional unit is assigned a weight corresponding to its success rate of execution.
This version of the algorithm is used to derive task trees for our task tree generation procedure. A limitation of the previous retrieval approach is that it cannot generate a task tree if the exact recipe does not exist in the graph. For example, it cannot produce the task plan of a ``corn soup" even though it knows about ``potato soup". Also, it cannot infer the task plan for a new ingredient that was never seen before.

\section{Task Tree Generation and Modification}
\label{sec:task_tree_mod}

In this section, we discuss our methodology to generate a task tree from any set of ingredients $\Ingredients$ and associated dish type $\DishType$ using FOON. $\Ingredients$ may contain some ingredients that are missing in FOON. To integrate those ingredients in the task tree, the proposed algorithm uses the concept of \textit{transfer learning}. The main idea is to find an existing recipe from FOON that closely matches $\Ingredients$ and modify it to produce a suitable final task tree. In the modification process, we make sure that the tree only contains the ingredients in $\Ingredients$ and with no extra objects. The whole process can be broken down into three main steps reviewed in the following subsections.

%%%%%%%%%%%%%%%%%%%%%%%%%%%%%%%%%%%%%%%%%%%%%%%%%%%%%%%%%%%%%%%%%%%%%%%%%%%%%%%%

\subsection{Identifying a Reference Goal Node}
\label{sec:task_tree_mod_A}
To generate a task tree, we first need to know what dish can be prepared with $\Ingredients$. For this purpose, we find a goal node in FOON that closely matches the required dish.
%The first major step is to find a goal node in FOON that closely matches the required dish, which is based on $\Ingredients$.
With this \textit{reference goal node} $\G$, a \textit{reference task tree} $\TaskTree$ could be extracted and then modified to suit the given ingredient set $\Ingredients$.
% In FOON, we extract the task tree of a recipe by searching with the final state of the dish, referred as goal node.
%
% That is why it is very important to identify the goal node first. 
%
%In this case, we try to find the goal node that closely matches to the required dish. %
To facilitate the selection of the ideal reference goal node, we created a recipe classification with 30 dish classes, such as \textit{Bread}, \textit{Soup}, \textit{Salad}, \textit{Pizza}, \textit{Drinks}, etc. 
FOON recipes (as subgraphs) were then categorized with these classes. 
%
% The purpose of creating this classification is that given a dish type, we want to find the similar recipes known to us. 
The purpose of this classification is to allow us to find similar recipes known to us, once given $\DishType$. 
Algorithm \ref{alg:finding_goal} uses this classification to find candidate goal nodes. For example, if $\DishType = salad$, candidates can be \{\textit{Greek Salad}, \textit{Potato Salad}, \textit{Caesar Salad}, ...\} depending on the salad recipes that exist in FOON. Each goal node stores a list of ingredients required to prepare that dish. Next, the algorithm checks each candidate goal node and selects the one that is most similar based on the ingredient set $\Ingredients$, drawing upon our previous work of applying similarity to identify similar objects and generalize knowledge in FOON~\cite{paulius2018functional}. The process for computing ingredient similarity is presented in Algorithm \ref{alg:compute_similarity}.  We use spaCy~\cite{spacy}, an open-source library for natural language processing, to get an embedded Word2Vec representation of each word. The Word2Vec model produces a 300-dimensional vector for each word and groups similar words together in the word embedding space. For instance, \textit{onion} and \textit{chives} will be close to each other but far away from \textit{fork}. Similarity is interpreted by a score ranging from 0 to 1, where 1 indicates that the two items are semantically the same. We consider two items similar if the score exceeds a predefined threshold value. Based on our experience, the threshold value is set to 0.90.     
 
\begin{algorithm}[t]
\caption{identify\_goal\_node}
\label{alg:finding_goal}
\begin{algorithmic}[1]
    \Statex \hspace{-2em} \textbf{Input:} Ingredients $\Ingredients$, Dish type $\DishType$
    \State $\Candidates$ $\gets$ Retrieve recipes of type $\DishType$ 
    \For{each $\Subgraph$ in $\Candidates$} 
        \State $\SetS \gets$ Ingredients of $\Subgraph$
		\State $\Score \gets$ Find similarity between $\Ingredients$ and $\SetS$
	\EndFor
	\State $\Goal \gets$ End product of subgraph with maximum $\mathsf{score}$
    \Statex \hspace{-2em} \textbf{Output:} $\Goal$
\end{algorithmic}
\end{algorithm}

%%%%%%%%%%%%%%%%%%%%%%%%%%%%%%%%%%%%%%%%%%%%%%%%%%%%%%%%%%%%%%%%%%%%%%%%%%%%%%%%

\begin{algorithm}[t]
\caption{compute\_similarity}
\label{alg:compute_similarity}
\begin{algorithmic}[1]
    \Statex \hspace{-2em} \textbf{Input:} Ingredient set $\Ingredients$ and $\SetS$  
    \State $\Count \gets 0$ 
    \For{each $i$ in $\Ingredients$}
        \For{each $j$ in $\SetS$}
        \State $\Score \gets $ Word2Vec similarity between $i$ and $j$
            \If{$\Score > \mathsf{Threshold}$}
                \State $\Count \gets \Count + 1 $
            \EndIf
        \EndFor
    \EndFor
    
    \Statex \hspace{-2em} \textbf{Output:} $\Count$
\end{algorithmic}
\end{algorithm}

%%%%%%%%%%%%%%%%%%%%%%%%%%%%%%%%%%%%%%%%%%%%%%%%%%%%%%%%%%%%%%%%%%%%%%%%%%%%%%%%

\subsection{Extracting a Reference Task Tree}

In this step, we extract the task tree of the reference goal node $\G$ and use it as a base of reference for our final task tree. In FOON, there can be various ways to reach to a goal, supporting that a dish can be prepared in possibly many ways. 
%We make the assumption that all preliminary items (i.e., in its raw or whole state) are available in the kitchen. 
We make the assumption that all preliminary items (i.e., in its raw or whole state) are available in the kitchen. 
The retrieval algorithm (shown as Algorithm \ref{alg:reference_retrieval}) evaluates all possible paths and selects the one with maximum overlap with $\Ingredients$. In a typically large FOON, exploring all paths can be time-consuming and memory inefficient. Hence, unlike the previous search algorithm~\cite{paulius2021weighted}, we choose a sub-optimal approach where some branches are pruned based on the ingredients stored at that node and their relevance with $\Ingredients$. 

% The retrieval procedure  checks for all possible paths and selects the one that has maximum overlapping with the given ingredients $\Ingredients$. In a very large FOON, exploring all paths can be time consuming and memory inefficient. Hence, unlike the previous search algorithm, we choose a sub-optimal approach where some branches are pruned based on the ingredients stored at that node and their relevance with $\Ingredients$. 

The algorithm works as follows: first, it finds all functional units $R$ where the goal node $\G$ is an output, which will serve as potential roots of the task tree. Initially, the task tree only contains the root functional unit. Next, we build the dependency tree for each root. For each input node, we search for the functional units $C^\prime$ that create them and have relevance with $\Ingredients$. For example, while searching for \textit{chopped onion}, we may find that \textit{onion} is not available in \textit{chopped} state, so we need \textit{whole onion} on a \textit{cutting board}. If \textit{whole onion} is available in the kitchen, we no longer need to explore that node. Additionally, if a unit in $C^\prime$ has an item not listed in $\Ingredients$, we remove it from our search queue and prune it from the unit. We compute the Cartesian Product $P$ of units in $C^\prime$ to generate valid paths with different combination of functional units. We refer the interested reader to \cite{paulius2021weighted} for the details of creating a tree $S$ from $P$. The search continues until all dependencies are met. We then perform DFS on $S$ starting from $R$ as the root of the tree to retrieve each individual path. 
Each path is a possible solution to reach the goal. 
As opposed to the procedure in \cite{paulius2021weighted}, where a path is selected that maximizes robot's success rate, we select the path with the most ingredient overlap with $\Ingredients$. If there are multiple paths with similar overlap, we select the shortest path, i.e., the one with minimum number of functional units. 

\begin{algorithm}[t]
\caption{retrieve\_reference\_task\_tree}
\label{alg:reference_retrieval}
\begin{algorithmic}[1]
    \Statex \hspace{-2em} \textbf{Input:} Desired object $\G$ and required ingredients $\Ingredients$ 
    \State Let $Q$ be the queue of nodes to search, $S$ be the list of tree nodes, and $R$ be the list of root nodes
    \State $R$ $\gets$ Find all functional units $\FU$ where $\G$ is an output
    \State $Q$.push($R$)
    \While{$Q$ is not empty}
        \State $L \gets$ $Q$.dequeue()
        \For{each functional unit $\FU$ in $L$}
        \For{each input $n$ in $\FU$}
            \State $C \gets$ Find all functional units that create $n$ 
            \State $C^\prime \gets$ Select units in $C$ that overlap with $\Ingredients$
            % \For{each $\FU$ in $C^\prime$}
            \State $E \gets$ Prune objects $x$ in all $\FU \in C^{\prime}$ if $x\notin\Ingredients$ 
            % \EndFor
            \State $P \gets$ Take Cartesian Product from $E$
            \Statex{\textit{\{Refer to \cite{paulius2021weighted} for the detail of tree node creation\}}}
            \State add $P$ to $S$ as a tree node
            \State $Q$.enqueue($P$)
        \EndFor
        \EndFor
    \EndWhile
    
    \State $\SetT \gets$ Use DFS on $S$ from $R$ to find all possible paths 
    
    % \Statex{\textit{\{Find path with maximum overlapping score with $\Ingredients$\}}}
    \For{each path $U$ in $\SetT$}
        \State score = compute\_similarity($\Ingredients$, $U$)
        \State $\TaskTree \gets$ Choose path that has maximum score
    \EndFor
    
    % \State $\TaskTree \gets$ 

    \Statex \hspace{-2em} \textbf{Output:} $\TaskTree$
\end{algorithmic}
\end{algorithm}

%%%%%%%%%%%%%%%%%%%%%%%%%%%%%%%%%%%%%%%%%%%%%%%%%%%%%%%%%%%%%%%%%%%%%%%%%%%%%%%%

\subsection{Task Tree Modification}

The reference task tree $\TaskTree$ gives a blueprint about how the tasks should be executed, but it may not contain all the required ingredients. Therefore, we need to modify $\TaskTree$ to incorporate missing ingredients. The modification procedure is shown in Algorithm \ref{alg:final_tree}. Each ingredient contains an object name and its state. An ingredient may exist in FOON in many different states. If an ingredient already exists in $\TaskTree$ in its required form, no adjustments are required. Otherwise, we have to search FOON for a task tree that makes that ingredient, which we refer to as a \textit{subtree}. It is possible that the ingredient does not exist in FOON, or it may exist in a different state. We resolve these cases by using the concept of semantic similarity (lines \ref{start_substitution}-\ref{end_substitution}). 

\textbf{Missing object type}: In our work, we make the assumption that objects of the same category can be manipulated in similar ways. We present each object name in a 300 dimensional word embedding space $E$ using its Word2Vec representation. Given an object $x$, we select $y$ as its equivalent ingredient if, among all ingredients, $y$ has the minimum distance from $x$ in $E$. Next, we retrieve the subtree of $y$ and substitute $y$ with $x$.
For example, if the required ingredient is \textit{sliced carrot} and we do not have \textit{carrot} in FOON, we look at how to make \textit{sliced cucumber} since \textit{carrot} and \textit{cucumber} are very close in the object embedding space. Afterward, we substitute \textit{carrot} with \textit{cucumber} in the subtree.

\textbf{Missing state}: Similar to the missing object case, if an state is missing, we can use the knowledge of how to achieve an equivalent state and apply it to reach a required state. To facilitate the process, we have created a state classification with 12 different categories, such as raw, finely separated, coarsely separated, liquid, etc., which are based on~\cite{jelodar2018identifying}. All states in FOON and Recipe1M+ are categorized using this state classification. With this mapping, we can use existing states as a blueprint for the other states in the same class. For example, if the required ingredient is \textit{sliced potato} and the state \textit{sliced} is unknown to us, we retrieve the task tree of an object with a similar state, such as \textit{diced potato}. We then substitute the similar state labels with the target state. While we substitute a state, we also have to choose the appropriate manipulation for which the state is valid. For instance, with the states \textit{diced} and \textit{sliced}, we treat the appropriate action with different motion verbs \textit{dice} and \textit{slice}. For this, we find the most frequently used motion verb associated with the required state and use it for the motion node label.  

\textbf{Integration of ingredients:} Once we retrieve a subtree and perform the necessary substitutions, the challenge is to connect it to the reference tree. To do this, we find functional units in $\TaskTree$ that can easily be modified, which are those that new ingredients can be added as input nodes. For example, any units that involve mixing or adding ingredients to a container can accept any number of inputs, whereas a robot cannot cut or chop multiple items at the same time. Therefore, for each missing ingredient, we connect the subtrees to those kinds of functional units, resulting in a connected acyclic graph. We assume that all the ingredients required to prepare the dish are given as input and no extra ingredients are needed. That is why in the post-processing phase, we remove all the ingredients from $\TaskTree$ that are not part of $\Ingredients$ (line \ref{removal}). Finally, $\TaskTree$ will have a list of functional units containing the input ingredients that, if executed sequentially, will produce the desired dish. Figure~\ref{fig:task_tree} shows an example of a generated task tree for a \textit{salad} recipe.

\begin{algorithm}[t]
\caption{construct\_final\_task\_tree}
\label{alg:final_tree}
\begin{algorithmic}[1]
    \Statex \hspace{-2em} \textbf{Input:} Ingredients $\Ingredients$, Dish type $\DishType$
    
    \State $\G \gets $ identify\_goal\_node($\Ingredients$)
    
    \State $\TaskTree \gets $ retrieve\_reference\_task\_tree($\G$) 

    \For{each ingredient $ing$ in $\Ingredients$}
        \If{$ing$ does not exist in $\TaskTree$}
            \If{$ing$ exists in $\FOON$}
                \State $\Subtree \gets$ Find task tree of $ing$
            \Else
                \State $eq \gets$ Find equivalent ingredient in $\FOON$ \label{start_substitution}
                \State $\Subtree \gets$ Find task tree of $eq$ object
                \State Substitute object, state and motion in $\Subtree$ \label{end_substitution}
            \EndIf
            
            Connect $\Subtree$ to $\TaskTree$
        \EndIf
        
    \EndFor
    
    \State Remove all ingredients $x$ from $\TaskTree$ if $x \notin \Ingredients$ \label{removal}

    \Statex \hspace{-2em} \textbf{Output:} $\TaskTree$
\end{algorithmic}
\end{algorithm}

%%%%%%%%%%%%%%%%%%%%%%%%%%%%%%%%%%%%%%%%%%%%%%%%%%%%%%%%%%%%%%%%%%%%%%%%%%%%%%%%

\section{Experiments}
\label{sec:experiments}

In this section, we present our experiments to test the flexibility of FOON-generated task trees. We also report the accuracy of our algorithm and analyze its performance.

\subsection{Dataset}
We use the Recipe1M+ dataset~\cite{marin2019learning} to create our dataset of recipes for our evaluation by selecting recipes for the dish classes introduced in Section~\ref{sec:task_tree_mod_A}. To make our evaluation more robust, we excluded dish classes for which we had less than 3 recipes in FOON. With too few recipes in a class, it makes it difficult to gain a general idea about the dishes of that class. Among the selected dish types, we selected 3 classes for our evaluation: {\textit{Salad}, \textit{Omelette}, \textit{Drinks}}. FOON has 10 \textit{Salad}, 7 \textit{Omelette}, and 8 \textit{Drinks} recipes. Next, we randomly selected 70 recipes for each category (for a total of 210 recipes) from Recipe1M+. Since Recipe1M+ does not have dish class assignments, we use the recipe titles to derive the category of a recipe. Table~\ref{table:statistics} shows a summary of the dataset used for our experiment. Most of these recipes require ingredients that do not exist in FOON, which makes this dataset suitable to evaluate the flexibility of our algorithm. 

\begin{table}[ht]
\centering
\caption{Overview of the dataset created from Recipe1M+}
\begin{tabular}{|p{0.20\linewidth} | p{0.20\linewidth} | p{0.20\linewidth}| p{0.20\linewidth}|}
\hline
\textbf{Dish Category} & \textbf{Total \# recipe in Recipe1M+} & \textbf{Avg \# ingredients per recipe} & \textbf{Avg \# unseen ingredients per recipe} \\ \hline
\textit{Salad} & 58,869 & 9 & 3 \\ \hline
\textit{Omelette} & 686 & 7 & 2 \\ \hline
\textit{Drinks} & 2,732 & 5 & 1 \\ \hline

\end{tabular}
\label{table:statistics}
\end{table}

\subsection{Evaluation Strategy}
Previously through a user study, we have shown that the FOON subgraphs are equivalent to Recipe1M+ recipes in terms of correctness, completeness, and clarity \cite{sakib2021evaluating}. This was our first effort to evaluate the quality of a task tree, as it is impossible to judge a recipe by only relying on automated testing such as Intersection over Union (IoU). This is because there can be many ways to prepare a meal. Furthermore, despite having the same ingredients, one recipe can differ from another recipe in terms of cooking steps while both are valid or correct. Hence, we chose to evaluate the FOON-generated task trees by manually checking the task plan for each ingredient. Although it requires a lot of manual effort, we believe that human judgement is necessary to evaluate any recipe. To make the process easier, instead of checking the functional units in the task tree, we check each ingredient by reviewing its \textit{progress line}. 

\begin{figure}[t]
	\centering
	\includegraphics[width=\columnwidth]{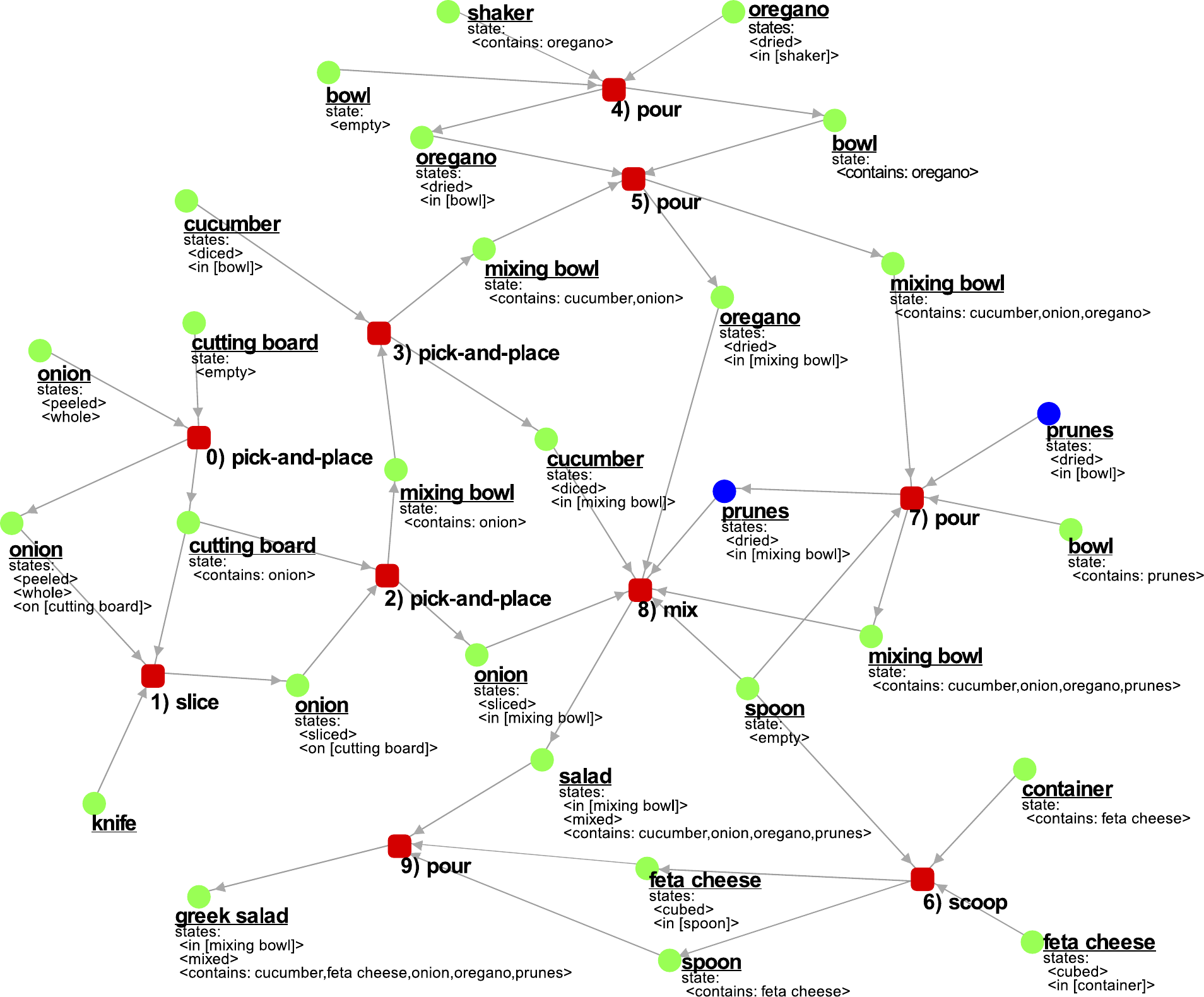}
	\caption{An example of a generated task tree for the input ingredients \{\textit{peeled onion}, \textit{diced cucumber}, \textit{dried prunes}, \textit{dried oregano}, \textit{dried feta cheese}\} (best viewed in color). The unseen ingredient added with 62.9\% confidence using the method of substitution is shown as blue nodes.}
	\label{fig:task_tree}
\end{figure}

\begin{figure}[t]
	\centering
	\includegraphics[width=\columnwidth]{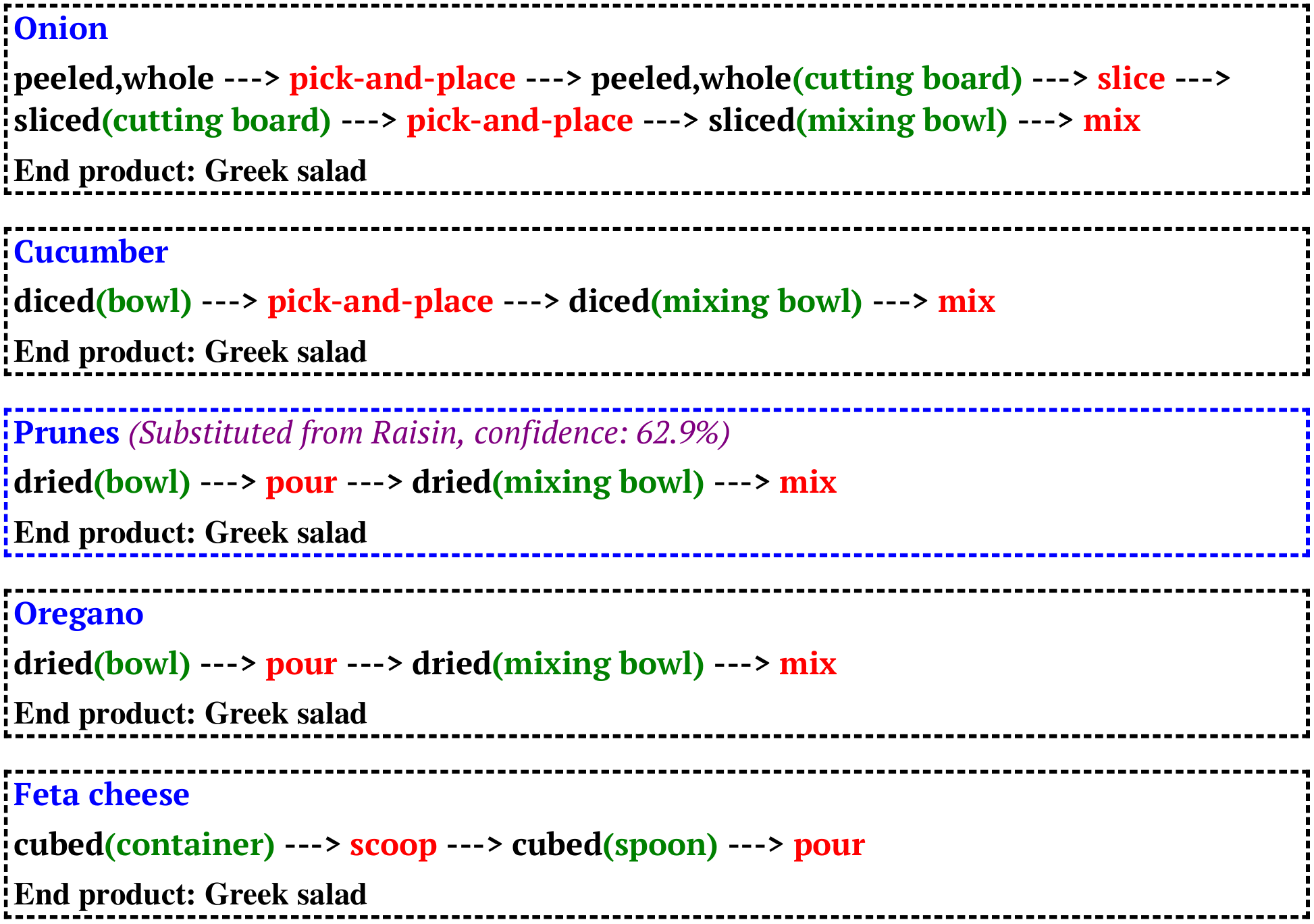}
	\caption{Illustration of state progression for each ingredient in the generated task tree in Figure \ref{fig:task_tree} (best viewed in color). The unknown ingredient \textit{prunes}, highlighted by a blue dashed box, is added with a state progression of 62.9\% confidence.}
	\label{fig:progress_line}
\end{figure}

\textbf{Progress Line:} In the cooking process, each ingredient can undergo several changes in its physical state (i.e., from \textit{whole} to \textit{sliced} and/or location (i.e., from \textit{cutting board} to \textit{bowl}). We refer to this sequence of changes as an ingredient's progress line; we can create this using the final task tree. We developed a tool to view the progress line of each ingredient and manually evaluate generated task trees. Figure~\ref{fig:progress_line} shows an example of progress lines generated from the task tree in Figure~\ref{fig:task_tree}, where objects, states and motions are colored in black, green and red respectively. Visualizing state transitions in this manner makes it easier to verify generation output.

% \begin{figure} 
%     \centering
%   \subfloat[A correct progress line of Leeks taken from an Omelette recipe.]{%
%       \includegraphics[width=0.48\textwidth]{figures/task_tree4.pdf}}
%     \hfill
%     \\
%   \subfloat[An incorrect progress line of Capers taken from a Salad recipe; The physical state of Capers can not be ``stem" like Parsley. ]{%
%         \includegraphics[width=0.48\textwidth]{figures/progress_line.pdf}}
    
%   \caption{Example of progress lines using for two ingredients that FOON has no knowledge about. It learns from equivalent ingredients and uses substitution method to create these progress lines.}
%   \label{fig:task_tree} 
% \end{figure}

\subsection{Results}
\begin{figure}[t]
	\centering
	\includegraphics[width=\columnwidth]{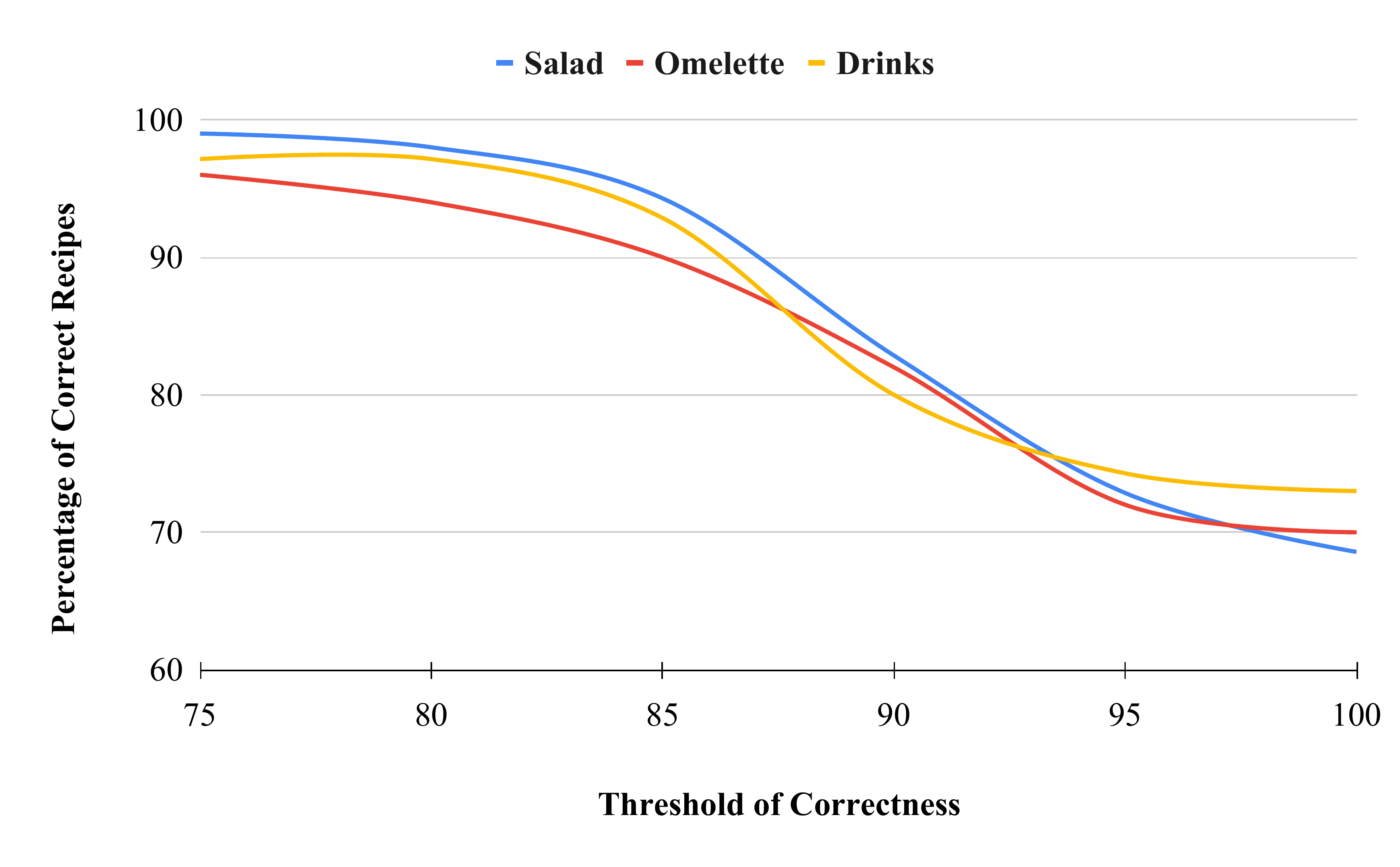}
	\caption{Graph showing the relationship between the threshold of correctness (X-axis) of each recipe and the number of recipes (Y-axis) we can generate with that threshold.} 
	\label{fig:result}
\end{figure}

Using the framework described in Section \ref{sec:task_tree_mod}, we generated the task trees for the selected 210 recipes. Each task tree was thoroughly evaluated using the progress line visualizer. When checking each recipe, we mark each of its ingredients as \textit{incorrect}, \textit{partially incorrect} or \textit{correct}. A label of {partially incorrect} means that the recipe has only one error in its state or motion labels.
We do this by giving each ingredient a score from \{0,1,2\}, where `0' means that the progress line is incorrect, `1' means partially incorrect, and `2' means correct. 
% Thus we identify how correct a generated recipe is.
In our experiments, we found that the \textit{Salad}, \textit{Omelette} and \textit{Drinks} recipes are \textbf{95.6\%}, \textbf{96.14\%} and \textbf{96.46\%} correct on average, respectively. We also varied the thresholds of correctness to see how many recipes we can generate with that threshold. In Figure~\ref{fig:result}, we report the percentage of correctly generated recipes with varying degrees of thresholds. For example, it shows that we can generate 94\% \textit{Salad}, 90\% \textit{Omelette} and 93\% \textit{Drinks} recipes if the threshold of correctness for each recipe is set to 85\%. When the threshold is 100\%, a task tree must have correct progress lines for each ingredient. In this case, we observed that the results are generally worse for \textit{Salad} compared to others, as \textit{Salad} recipes in Recipe1M+ typically have more unseen ingredients on average, which is supported by Table~\ref{table:statistics}.

\begin{figure}[t]
	\centering
	\includegraphics[width=8cm]{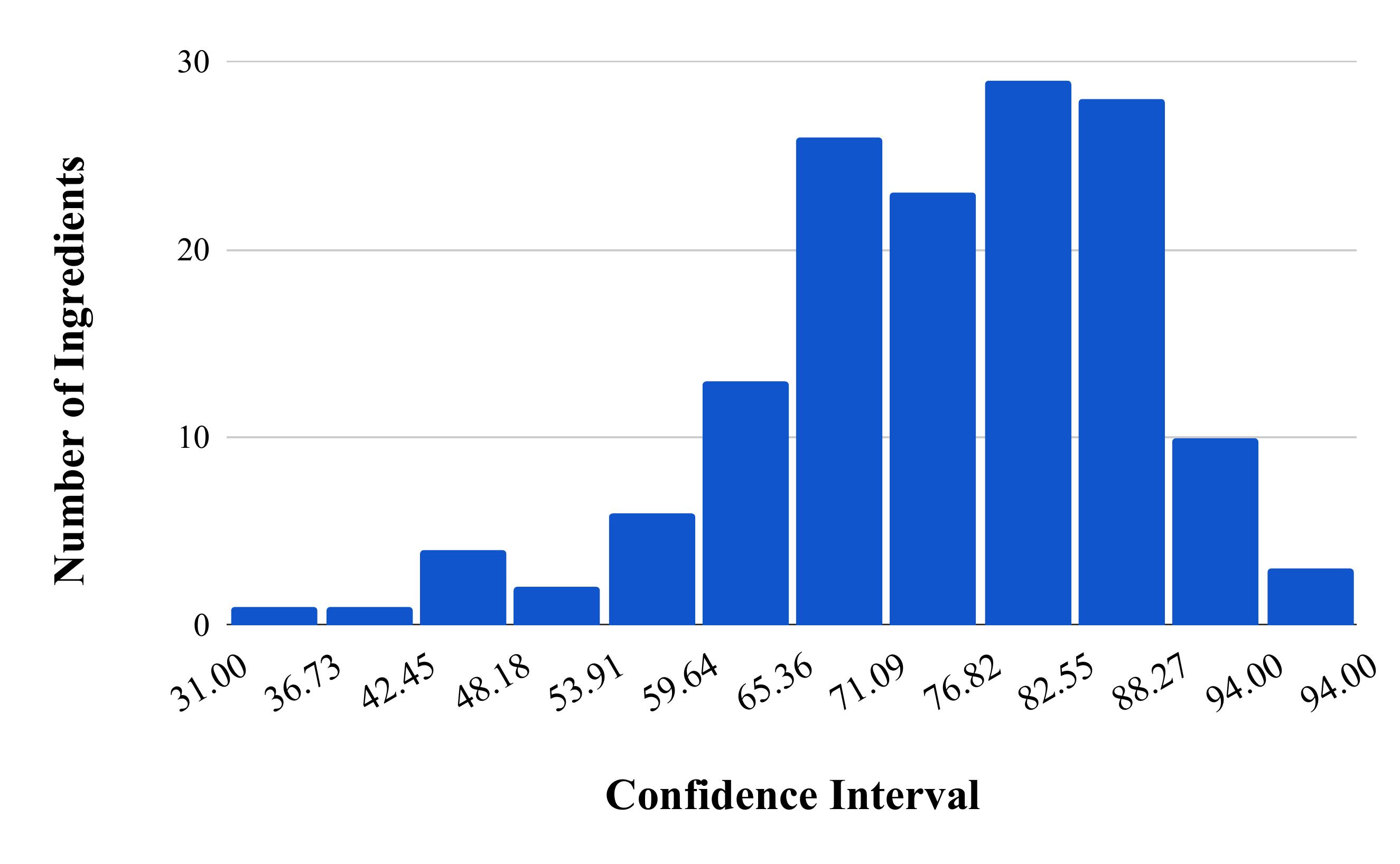}
	\caption{Histogram of ingredient similarity confidence scores for unseen ingredients in the dataset.}
	\label{fig:histogram}
\end{figure}

\begin{figure} 
    \centering
    \subfloat[A correct progress line of \textit{leeks} taken from an \textit{Omelette} recipe.]{
    \includegraphics[width=0.48\textwidth]{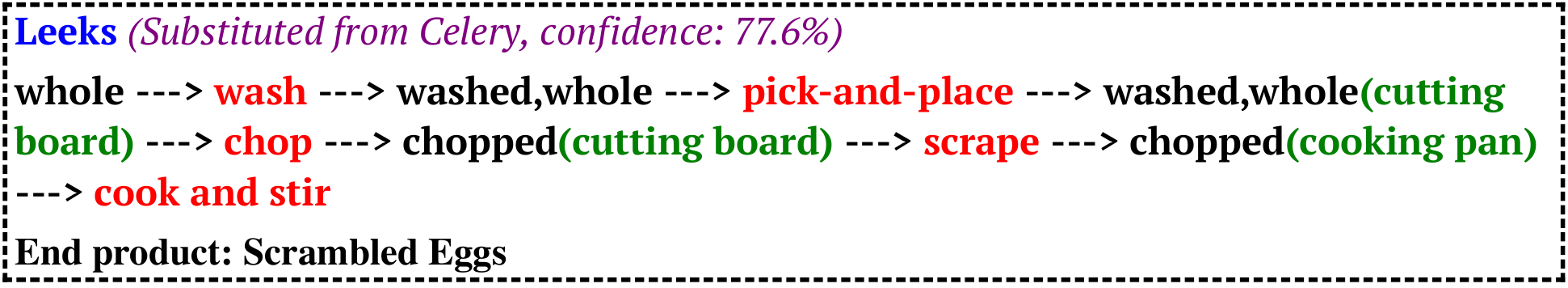}}
    \hfill
    \\
    \subfloat[An incorrect progress line of \textit{capers} taken from a generated \textit{Salad} recipe; \textit{capers} cannot be in a \textit{stem} state like \textit{parsley}.]{
    \includegraphics[width=0.48\textwidth]{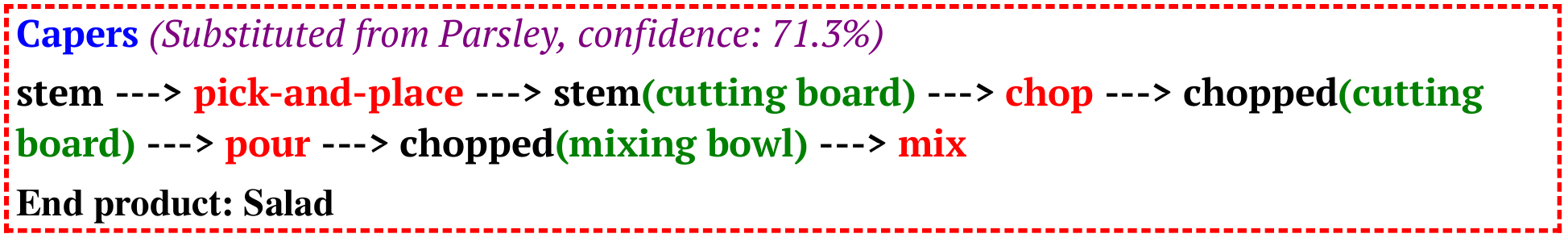}}
    \caption{Example of progress lines for two ingredients that FOON has no knowledge about. It learns how to process them from equivalent ingredients and uses the substitution method to infer about the correct chain of state changes.}
    \label{fig:substitution} 
\end{figure}

\textbf{Impact of substitution}
Recipe1M+ recipes consist of many ingredients that do not exist in FOON. The proposed algorithm in Section~\ref{sec:task_tree_mod} uses method of substitution to handle such ingredients. For an ingredient, if we have a very close equivalent, we can perform substitution with high confidence. In most cases, this will generate a correct progress line. However, a substitution with low confidence often leads to inaccuracy. Figure~\ref{fig:histogram} shows a histogram of our confidence of substitution for the unseen ingredients in our dataset. The confidence for an ingredient is measured by the similarity to its closest object in the word embedding space and is represented as percentage. Although we can correctly integrate many unseen ingredients into our task tree via substitution, this part is still the most critical section of recipe generation process. 
%We have observed that substitution is the reason of 80\% errors in all of the incorrect progress lines. 
Examples of progress lines generated using substitution method are shown in Figure~\ref{fig:substitution}. In this figure, \textit{capers} has an incorrect progress line because the most equivalent ingredient of \textit{capers} in FOON is \textit{parsley}, according to our Word2Vec model. However, they have completely different physical shapes and properties; hence, they would require different methods to process them. On the other hand, \textit{leeks} and \textit{celery} have the same physical structure. Therefore, the progress line of \textit{leeks} accurately reflects the changes of physical state and location after each action.

% \subsection{Discussion}
One way to improve task tree accuracy is by enriching FOON through the addition of new recipes with diverse set of ingredients. In this way, our algorithm will be better suited to finding ideal candidates for equivalent ingredients, and consequently, errors caused by inaccurate substitution will be reduced. For instance, in our experiments, we found that the errors in the progress line of \textit{capers} in Figure~\ref{fig:substitution} occurred due to the lack of a suitable \textit{olive}-like object in FOON. To verify this, we added \textit{olives} to one recipe in FOON and generated the task tree again, after which we were able to generate the correct progress line of \textit{capers} with 86.8\% confidence. 
% One of the limitations of FOON is that it can not generate a task plan for two dishes at one recipe. For instance,  

\subsection{Analysis of the Algorithm}
The time and space complexity of our search algorithm is dominated by the depth-first search in the reference task tree retrieval process. FOON grows as new recipes are added to it; as a result, the search creates more possible task trees by exploring new paths. Additionally, we have to store all the task trees so that at the end of the search, we can choose the best one. Since we already have information about the required ingredients, we can prune branches for irrelevant ingredients during the search process so we can save time and memory. We can further improve the searching time by storing the previously retrieved reference task trees, so that we can quickly fetch in them for future queries.

If all required ingredients are available in the reference task tree, we do not need to do any substitution or integration of ingredients. In that case, it will always provide the optimal solution in a limited time. We define optimality by the number of actions required to perform.   
A task tree is optimal if it has the minimum possible functional units. Optimality cannot be guaranteed when there are unseen ingredients.    

% One of the limitations of FOON is that it can generate task tree for one recipe at a time. But, sometimes we observe dishes that is a combination of different recipes such as  

\section{Conclusion and Future Work}
\label{sec:con}
In summary, we introduced a \textit{task tree generation} pipeline that will allow a robot to obtain task plans that creatively applies existing knowledge to novel problems, which possibly include never-before-seen ingredient combinations, using the concept of semantic similarity.
This idea is motivated by a human's ability to modify plans or recipes dynamically.
With our methodology, task trees were generated using knowledge from Recipe1M+, and in our experiments, we showed that these plans can be generated with minimal error.

It is obvious that adding more recipes to the knowledge base will help to adapt to novel scenarios in a better way. We aim to reduce the manual effort of subgraph creation by extracting information from a video/textual recipe using a machine learning approach. Moreover, to concretely evaluate our method, we will compare generated task trees with the ground-truth recipes from Recipe1M+ using both manual checking and IoU computation~\cite{sakib2021evaluating}. 
We will also explore how a real robot can execute such generated task trees.

\section*{Acknowledgement}
\noindent This material is based upon work supported by the National Science Foundation under Grant Nos. 1910040 and 1812933.

\bibliographystyle{unsrt}
\bibliography{ref}

\end{document}